\documentclass{article}

\usepackage[preprint]{neurips_2026}

\usepackage[utf8]{inputenc}
\usepackage[T1]{fontenc}
\usepackage{hyperref}
\usepackage{url}
\usepackage{booktabs}
\usepackage{amsfonts}
\usepackage{amssymb}
\usepackage{amsmath}
\usepackage{amsthm}
\usepackage{enumitem}
\usepackage{nicefrac}
\usepackage{microtype}
\usepackage{xcolor}
\usepackage{graphicx}

\newtheoremstyle{taeplain}{1ex plus .25ex}{1ex plus .25ex}{\itshape}{}{\bfseries}{.}{ }{}
\newtheoremstyle{taedefn}{1ex plus .25ex}{1ex plus .25ex}{}{}{\bfseries}{.}{ }{}
\theoremstyle{taeplain}
\newtheorem{proposition}{Proposition}
\newtheorem{corollary}{Corollary}
\theoremstyle{taedefn}
\newtheorem{definition}{Definition}
\newtheorem{postulate}{Postulate}
\newtheorem{assumption}[definition]{Assumption}

\newcommand{\teArrow}{\ensuremath{\rightarrow}}

\newcommand{\teApprox}{\ensuremath{\approx}}

\title{The Architecture of Errors:\\From Universal Impossibility to Patch-Local LLM Reliability}

\author{%
  Mikhail L. Arbuzov\\
  Independent Researcher\\
  \texttt{mike.arbuzov54@gmail.com}\\
  \And
  Lee Mosbacker\\
  Independent Researcher\\
  \texttt{lee.mosbacker@gmail.com}\\
  \And
  Sisong Bei\\
  Independent Researcher\\
  \texttt{qurining@gmail.com}\\
  \AND
  Ziwei Dong\\
  Independent Researcher\\
  \texttt{ziwei.dong@alumni.emory.edu}\\
  \And
  Dmitri Kalaev\\
  Independent Researcher\\
  \texttt{kalaevdr@gmail.com}\\
  \And
  Alexey Shvets\\
  Palo Alto Networks\\
  \texttt{ashvets@paloaltonetworks.com}\\
}

\begin{document}

\maketitle

\begin{abstract}
Universal LLM reliability is not a finite-library problem: across all possible tasks, tools, schemas, knowledge sources, and evaluator expectations, new intervention-distinguishable failure modes can appear without bound, so no finite intervention dictionary can guarantee bounded residual error for every such mode. But deployed systems do not operate over the whole universe. They operate inside operationally bounded patches (legal review, medical RAG, code repair, customer-support agents, contract extraction) with recurring tasks, schemas, tools, and evaluator expectations. Within such patches, empirical evidence suggests failures are sparse, repetitive, and concentrated in a small recurring catalogue, so reliability becomes a local catalogue-discovery and intervention-coverage problem rather than an exponential token-length problem. We formalize this transition with two propositions and one corollary. Proposition~\ref{prop:nounidict} is the worst-case-mode-wise negative result: no finite intervention dictionary covers every distinguishable failure mode of an unbounded domain. Corollary~\ref{cor:invdiscovery} is the inverse-discovery implication: the logarithmic upper bound on mode discovery cannot accommodate linearly more distinct tail modes without exponentially more observed hard-failure events. Proposition~\ref{prop:budget} is the positive patch-local result: under log active-mode exposure and head-heavy coverage, a sufficient per-hard-decision intervention budget grows polylogarithmically in sequence length and becomes domain-constant once the patch catalogue saturates. The framework relocates rather than dissolves long-context difficulty: where the number of hard decisions itself grows with task length, reliability remains hard; the contribution is to identify the on-axis intervention rather than to make those regimes easy.
\end{abstract}

\section{Introduction}
\label{sec:intro}

The standard worry about long-context autoregressive generation is exponential. If every token has independent error probability $e$, then after $n$ tokens the chance of a fully correct output is $(1-e)^n$, which collapses to zero for any nontrivial $e$ and large enough $n$~\citep{lecun2023doomed,dziri2023faith}. Earlier work in this series~\citep{arbuzov2025beyond} argued that this worry is misplaced because errors are not distributed uniformly across tokens: only $5$--$10\%$ of tokens are ``key'' (genuinely dependent on long-range context~\citep{fang2025perplexity}), while the remaining majority become nearly deterministic once enough context accumulates. Replacing $(1-e)^n$ with the two-rate model $P(\text{correct}) = (1-e_{\text{key}})^{k} (1-e_{\text{non}})^{n-k}$ with $k \ll n$ scaling sublinearly in $n$ recovers the observed long-context coherence and converts the reliability question from ``how does $n$ grow?'' to ``how does $k$ grow?''.

\paragraph{From sparse tokens to recurring patterns.}
This paper takes the next step. Sparsity tells us \emph{where} errors live; the natural follow-up is \emph{what} they are. Recent failure-mode atlases supply a striking empirical answer: errors are not only sparse, they are also \emph{repetitive}. ErrorAtlas~\citep{ashurytahan2026erroratlas}, covering 83 models across 35 datasets, organises observed failures into 17 named categories sorted by prevalence, with a long-tailed distribution heavily concentrated in the head. In code, two error types (\texttt{AssertionError} and \texttt{NameError}) cover 86.35\% of failures on HumanEval across 14 LLMs, replicated across 23 models on HumanEval Pro and MBPP Pro~\citep{wen2024fixing}. In math, MWPES-300K~\citep{sun2025mwpes} categorises 304,865 errors from 15 LLMs across four math word-problem datasets and reports that dataset characteristics shape error patterns systematically. The same picture appears in multi-hop QA~\citep{zhang2026weakestlink}, agentic tool use~\citep{cemri2025mast}, and retrieval-augmented generation~\citep{wood2024acurai}. These observations suggest a third layer in the reliability architecture, complementing key-token sparsity (Layer 1) and within-key stratified-manifold structure (Layer 2 in~\citealt{arbuzov2025beyond}). Within the small set of key tokens, only a fraction $\beta$ produce \emph{hard} failures, and inside bounded deployment patches those hard failures appear to cluster into a finite or effectively capped catalogue of recurring modes whose size grows much more slowly than the number of observed events.

\paragraph{Contributions.}
The central contribution is a shift in the reliability object. Universal LLM reliability is not a finite-library problem, but patch-local reliability can be treated as catalogue discovery and intervention coverage. We formalise this transition with two propositions and one corollary.

Proposition~\ref{prop:nounidict} is the worst-case-mode-wise negative result: if an unbounded domain keeps producing intervention-distinguishable failures, no finite intervention dictionary can guarantee bounded residual error for every mode of the domain. Corollary~\ref{cor:invdiscovery} is the inverse-discovery implication of the logarithmic upper bound: the bound cannot accommodate linearly more distinct tail modes without exponentially more observed hard-failure events, so open-domain tail discovery has rapidly diminishing returns. Proposition~\ref{prop:budget} is the matching positive result inside a fixed deployment patch. Under log active-mode exposure and head-heavy coverage, a sufficient per-hard-decision intervention budget satisfies $m \geq \lceil |C_{\text{eff}}|^{1 - \varepsilon / e_{\text{hard}}} \rceil$, with a doubly-logarithmic sequence-length rate as the optimistic pre-cap special case and a domain-constant budget once the patch catalogue saturates. The sequence-level analogue is strictly tighter and approaches full-catalogue coverage as $k$ grows; we surface that asymmetry rather than bury it.

Around this transition the paper does four supporting things. A three-layer framework, comprising sparsity ($\alpha$), hard-token stratification ($\beta$), and patch-local mode catalogue ($|C_D|$), separates \emph{where} errors occur, \emph{what} recurring forms they take, and \emph{which} capability interventions address them (\S\ref{sec:framework}). Logarithmic mode discovery is stated as an empirical postulate, not a theorem, calibrated against ErrorAtlas, HumanEval, and MWPES with $\sigma \in [0.87, 1.85]$ across anchors and $\sigma \approx 1.85$ carried as a conservative planning value. Section~\ref{sec:evidence} synthesises evidence for failure clustering, cluster-selective interventions, and sublinear length scaling, drawing on $\sim$60 prior published results including the six-axis capability-elimination harvest of 28 quantitatively-anchored citations stratified into Patterns A/B/C (Appendix~\ref{app:taxonomy}). And the most-cited steep-decay counter-evidence (Appendix~\ref{app:reaudits}) is re-audited and shown to decay primarily over task-structure variables, including compositional graph size, fact count, log-time horizon, capacity threshold, and evidence scope, rather than raw token length.

\section{Related Work}
\label{sec:related}

\paragraph{Failure-mode taxonomies.}
ErrorAtlas~\citep{ashurytahan2026erroratlas}, MWPES-300K~\citep{sun2025mwpes}, the HumanEval categorisation of~\citet{wen2024fixing}, and the RFMDataset~\citep{guo2025rfmdataset} together suggest that, at any given corpus size, the number of distinct named failure modes is small (typically 8--20) and the cumulative coverage of the top modes is high. Domain-specific taxonomies for multi-hop QA~\citep{zhang2026weakestlink}, multi-agent systems~\citep{cemri2025mast}, and tool-augmented agents~\citep{yao2024taubench} report the same pattern.

\paragraph{Targeted interventions.}
Each named cluster has been the subject of focused intervention research. Python execution closes most of the arithmetic-error cluster~\citep{gao2022pal,chen2022pot,gou2024tora}; constrained decoding eliminates format violations~\citep{suresh2025dingo,wang2025slot,dong2024xgrammar}; execution feedback resolves most code-logic errors~\citep{shinn2023reflexion,huang2023agentcoder}; process reward models reduce step-level reasoning errors~\citep{wangp2024mathshepherd,lightman2024verify}; RAG strongly reduces hallucinations~\citep{wood2024acurai}; preference optimisation curbs over-refusal~\citep{karaman2024porover}; structured uncertainty fixes most tool-call failures~\citep{suri2025clarification}. Two structural observations recur: each intervention is \emph{cluster-selective} (residuals belong to a different named cluster, not to the targeted one), and \emph{additivity is approximate but not perfect}~\citep{patel2026sixsigma,le2026interaction}.

\paragraph{Length-decay benchmarks.}
A parallel literature measures the \emph{shape} of the reliability decay curve. Mild-decay results (Loong~\citep{wangm2024loong}, GSM-$\infty$~\citep{zhou2025gsminfinite}, RULER~\citep{hsieh2024ruler}, anchor-based LLMs~\citep{pang2024anchor}) report log-linear, sigmoidal, or threshold-like decay inconsistent with smooth $(1-\varepsilon)^n$. Steep-decay results~\citep{dziri2023faith,kuratov2024babilong,kwa2025horizon,wan2026fano} are sometimes cited as exponential-compounding evidence. The apparent tension is reduced on careful reading: each of the steep-decay results we re-audit decays over a variable distinct from raw token length, which Appendix~\ref{app:reaudits} documents.

\paragraph{Gap.}
What is missing is a quantitative bridge between the small, repeating taxonomy catalogue and a polylog intervention budget. We provide that bridge in \S\ref{sec:framework}.

\section{Theoretical Framework}
\label{sec:framework}

\paragraph{Four levels of $|C|$.}
Before any proposition, a definitional setup. LLM error analysis routinely conflates four objects that the rest of this paper needs to keep apart:

\begin{itemize}[leftmargin=1.8em,itemsep=0.3ex]
\item \textbf{L1: failure events.} Concrete LLM errors observed in a benchmark; the raw data behind any taxonomy.
\item \textbf{L2: empirical taxonomy categories.} Researcher-chosen labels grouping L1 events. ErrorAtlas's 17 categories, MWPES's top-4 classes, and HumanEval's \texttt{AssertionError}/\texttt{NameError} partition all live here. L2 is observable but depends on the taxonomer's resolution choices.
\item \textbf{L3: latent failure modes.} The unobserved underlying clusters that L2 approximates. L3 is the quantity Postulate~\ref{post:logmode} is morally about, but it cannot be measured directly; we treat L2 as a noisy proxy for L3.
\item \textbf{L4: capability axes / interventions.} The engineering unit: a Python interpreter, a constrained decoder, a retrieval-augmented generator. L4 is coarser than L2 in practice; one capability axis typically targets several L2 categories at once (\S\ref{sec:evidence}, Claim~B).
\end{itemize}

Throughout this section, $|C|$ refers to the \textbf{L2} count: that is what published taxonomies report. Postulate~\ref{post:logmode} is therefore \emph{engineering}-relevant because L4 is coarser than L2 (so a small intervention library can sweep across many categories at once), and \emph{epistemically} conditional because L2 is a noisy proxy for the actual L3 mode catalogue whose faithfulness has not been independently measured. Tagging the formal claims at the right level keeps the framework honest where casual error-talk slips.

\paragraph{Roadmap.}
The framework separates three questions that are often conflated. First, \emph{where} do errors occur? Only a sparse subset of key decisions. Second, \emph{what} kinds of errors recur? A local catalogue of failure modes. Third, \emph{what fixes them}? A smaller library of capability interventions. The two propositions of \S\ref{sec:proposition} formalise the transition from universal impossibility to patch-local tractability: there is no finite intervention dictionary that covers all possible deployments, but inside a fixed deployment patch a sufficient library is small and slowly growing.

\subsection{$\beta$-stratification of key tokens}

The two-rate model of~\citet{arbuzov2025beyond} distinguishes $k$ key tokens (error rate $e_{\text{key}}$) from $n - k$ non-key tokens ($e_{\text{non}} \ll e_{\text{key}}$). Empirical atlases suggest that even within the key-token class errors are not uniformly distributed: most key tokens are ``decisions'' for which the model has stable representations; only a fraction concentrate the actual failures.

\begin{definition}[Hard fraction]\label{def:hard}
Partition the $k$ key tokens of a sequence into easy and hard subsets,
\[
k_{\text{hard}} = \beta k, \qquad k_{\text{easy}} = (1-\beta) k, \qquad \beta \in (0,1),
\]
where hard key tokens have an elevated error rate $e_{\text{hard}}$ corresponding to manifold-transition decisions in the sense of~\citet[\S3.2]{arbuzov2025beyond}, and easy key tokens have $e_{\text{easy}} \teApprox e_{\text{non}}$.
\end{definition}

Under Definition~\ref{def:hard}, the composed sequence-level reliability becomes
\begin{equation}\label{eq:three-rate}
P(\text{correct}) \;=\; (1 - e_{\text{hard}})^{\beta k}\,(1 - e_{\text{easy}})^{(1 - \beta) k}\,(1 - e_{\text{non}})^{n - k}.
\end{equation}
We do not assume iid token errors: the three rates summarise \emph{conditional per-decision failure hazards} after conditioning on prior decisions being correct, and grouping by stratum yields the multiplicative survival expression. Because $e_{\text{easy}} \teApprox e_{\text{non}}$ once context accumulates, the failure rate is dominated by the $\beta k$ hard tokens, and we treat $e_{\text{hard}}$ as the load-bearing quantity. The parameter $\beta$ is \emph{latent}: failure-mode atlases report $\Pr(\text{category} \mid \text{error occurred})$, which does not identify $\Pr(\text{hard} \mid \text{key-token decision})$. We carry $\beta$ symbolically.

\subsection{Empirical postulate: logarithmic mode discovery}
\label{sec:postulate}

Stratification names a target, the $\beta k$ hard-token decisions, but does not yet tell us how many distinct ways those hard decisions can fail. The empirical answer from Section~\ref{sec:intro} is that failures repeat: a small number of recurring patterns covers most of the mass. The next question is how the catalogue's size grows with the number of observed failures, because that is the quantity an intervention library has to keep up with.

Two candidates exist in the type--token literature: Heaps' law (power-law, $|C| \approx K \cdot k_{\text{hard}}^{b}$ with $b \in [0.4, 0.6]$~\citep{manning2008ir}), and logarithmic discovery. \textbf{Zipfian rank-frequency does not imply logarithmic growth.} Heaps' law is the correct type--token consequence of Zipf-distributed events, and it is power-law, not logarithmic. We therefore state logarithmic mode discovery as an empirical postulate, defensible by direct measurement of error taxonomies, not as a theorem.

Two sample-size variables matter, and they are easy to conflate. Let $h(n) = \beta k(n)$ denote the number of hard-token decisions in a single sequence of length $n$; let $T$ denote the number of observed hard-failure events in a corpus used to discover the domain catalogue. These are not the same object: $h(n)$ controls per-sequence exposure, $T$ controls empirical discovery. We write $C_D$ for the full reachable catalogue inside domain patch $D$, $C_{\text{seen},D}(T)$ for the subset discovered after $T$ sampled failures, and $C_{\text{active},D}(n)$ for the subset a single sequence of length $n$ can activate. We write $\ln T$ and $\ln h(n)$ for clarity; small-sample bounds may be read with $\ln(1+T)$ and $\ln(1+h(n))$, and discrete counts such as $\beta k$ are interpreted either as expected values or as $\lfloor \beta k \rfloor$ asymptotically. These reformulations do not affect any rate claim below.

\begin{postulate}[Patch-indexed catalogue discovery]\label{post:logmode}
Within a fixed application domain $D$, the number of named recurring failure modes discovered after $T$ sampled hard-failure events is bounded by
\[
|C_{\text{seen},D}(T)| \;\leq\; \min\bigl(A_D + \sigma_D \ln T,\ |C_D|\bigr), \qquad \sigma_D > 0,\ A_D \geq 0.
\]
The cap $|C_D|$ is the domain-imposed ceiling. The postulate concerns catalogue \emph{discovery}, not the number of hard decisions inside a single sequence.
\end{postulate}

\begin{assumption}[Per-sequence active-mode exposure]\label{ass:active}
For a single sequence of length $n$ in domain $D$,
\[
|C_{\text{active},D}(n)| \;\leq\; \min\bigl(A'_D + \sigma'_D \ln h(n),\ |C_D|\bigr).
\]
Primed constants are distinct from those of Postulate~\ref{post:logmode}: corpus discovery and per-sequence activation need not share the same rate.
\end{assumption}

The two bounds answer different questions, and the rest of the paper picks the right one at each point. Proposition~\ref{prop:budget}'s sequence-length scaling rate is governed by $C_{\text{active},D}(n)$ via Assumption~\ref{ass:active}: how much of the catalogue can a single sequence touch? Full domain-library budgeting, the practical engineering target for a deployed system, uses $C_D$: how large does the library need to be to cover the domain's reachable failure modes? The cap $|C_D|$ enters both: once enough has been sampled (or activated) to saturate the domain ceiling, further growth stops and the budget becomes a domain-constant function of $|C_D|$ alone.

\paragraph{Empirical calibration.}
Existing LLM error taxonomies provide endpoint anchors for small named catalogues at observed corpus scales. ErrorAtlas, HumanEval-style code taxonomies~\citep{wen2024fixing}, and MWPES~\citep{sun2025mwpes} place $|C|$ roughly in the $8$--$20$ range across corpora ranging from approximately $10^4$ to $3 \times 10^5$ observed failures, yielding $\sigma \in [0.87, 1.85]$ under the simple $A=0$ logarithmic calibration. We use $\sigma \teApprox 1.85$ as a conservative planning value, not as a fitted law. These endpoint counts are not discovery curves: the missing test is a subsample-vs-discovered-modes measurement within a fixed deployment patch. Full calibration details are in Appendix~\ref{app:calibration}.

\paragraph{Capabilities are coarser than error categories.}
The category count $|C|$ is conservative for engineering because deployed interventions operate at the capability level, not the label level. A Python interpreter can remove execution-error components across arithmetic, unit conversion, counting, list manipulation, and date arithmetic; a constrained decoder can eliminate several structural output failures at once. The formal coverage problem is therefore closer to weighted set cover over failure mass than one-intervention-per-category counting. The ranked-category model used in Proposition~\ref{prop:budget} is a tractable proxy; the full capability harvest appears in Appendix~\ref{app:taxonomy}.

\paragraph{L2 $\to$ L4 as weighted set cover.}
Formally, let each L4 intervention $I_j$ cover a subset $S_j \subseteq C_D$ of the L2 mode catalogue, with per-mode residual-reduction weight $r_{ij} \in [0, 1]$ giving the fraction of mode $i$'s hard-error mass that $I_j$ removes. The deployment problem chooses a library $\mathcal{I}$ of size at most $m$ to maximise covered hard-error mass:
\[
\max_{\mathcal{I} : |\mathcal{I}| \leq m}\; \sum_{i \in C_D} p_i \, \max_{j : I_j \in \mathcal{I}} r_{ij}.
\]
Proposition~\ref{prop:budget} below uses the analytically tractable binary special case $r_{ij} \in \{0, 1\}$ in which the $m$-th-ranked intervention covers the $m$-th-ranked mode (one mode per intervention, ordered by marginal covered mass). In real deployments one L4 capability typically covers several L2 categories simultaneously, so the ranked-category budget of Proposition~\ref{prop:budget} is a conservative proxy for the true set-cover optimum: the empirically achieved library size for a given residual target may be a constant factor smaller than the formula predicts. Overlap, additivity, and prompt-channel interference are treated empirically in the capability-harvest of Appendix~\ref{app:taxonomy} (``Additivity and its limits''), not as a formal set-cover optimisation.

\paragraph{Operational patch.}
A deployment patch $D$ is more than a topic label. It is an operational tuple fixing, over a chosen time window, the components that determine which failure modes are reachable at all:
\[
D = (\mathcal{X},\, \mathcal{S},\, \mathcal{U},\, \mathcal{R},\, \mathcal{E},\, \mathcal{P},\, H,\, \tau),
\]
where $\mathcal{X}$ is the task input distribution, $\mathcal{S}$ the input/output schema family, $\mathcal{U}$ the user or client class, $\mathcal{R}$ the retrieval corpus or knowledge source, $\mathcal{E}$ the evaluator or acceptance criterion, $\mathcal{P}$ the policy and safety constraints, $H$ the workflow horizon, and $\tau$ the time window over which these components are fixed. A \emph{patch shift} occurs when one or more of these components changes enough to alter the reachable failure catalogue $C_D$. Legal review on contracts from a fixed jurisdiction is a different patch from legal review across mixed jurisdictions; medical RAG over a curated knowledge source is a different patch from medical RAG over arbitrary web content. The patch-local claims in this paper apply within a single, operationally fixed tuple of this form, not to ``the domain'' in any looser sense.

\paragraph{Domain patches cap the engineering problem.}
A fixed deployment patch restricts the task family, schemas, tools, workflows, and evaluator expectations. Prior work on localized representations~\citep{park2024linear,li2025stratified}, cross-domain performance variation~\citep{wangy2024mmlupro}, and long-tail knowledge~\citep{mallen2023popqa,kandpal2023longtail} supports the weaker claim that model behaviour is strongly domain-dependent; it does not directly measure $|C_D|$. We therefore treat the patch's reachable catalogue as finite or effectively capped as a modelling assumption motivated by domain heterogeneity, not as a theorem derived from boundedness. The full evidence breakdown is in Appendix~\ref{app:patch-evidence}.

\begin{corollary}[Inverse Discovery Cost]\label{cor:invdiscovery}
Postulate~\ref{post:logmode} caps the discovered catalogue from above. Reading the bound in the inverse direction, the corpus must contain at least
\begin{equation}\label{eq:invcost}
T \;\geq\; \exp\!\left(\frac{q - A_D}{\sigma_D}\right)
\end{equation}
observed hard failures before that cap can accommodate $q$ distinct discovered modes. Equivalently, the minimum sample budget compatible with the bound scales as $T_{\min}(q) = \exp((q - A_D)/\sigma_D)$, so accommodating $\Delta q$ further modes raises that minimum budget by a factor $\exp(\Delta q / \sigma_D)$.
\end{corollary}

\paragraph{Tightness reading.}
If the empirical discovery curve is approximately tight against the bound at the observed corpus scales, the inequality becomes an approximate equality, $T(q) \teApprox \exp((q - A_D)/\sigma_D)$. At the conservative calibration $\sigma_D \teApprox 1.85$, five additional discovered modes cost roughly $15\times$ more failures, and ten cost roughly $220\times$. The lower bound \eqref{eq:invcost} holds unconditionally; the multiplicative-cost reading needs tightness.

\paragraph{Engineering meaning.}
The claim is about \emph{newly distinguishable} tail modes, not ordinary failures, which remain common inside already-discovered modes. Combined with the head-heavy coverage model of \S\ref{sec:coverage}, the picture is one of asymmetric returns: the head is cheap to find and absorbs most error mass; the tail is expensive to discover and removes less residual error per mode. Full derivation, the Heaps polynomial variant, the saturation regime, and a mode-mediated capability-gain sub-corollary are in Appendix~\ref{app:invdiscovery}.

\subsection{Coverage by a targeted intervention library}
\label{sec:coverage}

With the catalogue defined, the next question is how much residual error a library of $m$ targeted interventions actually closes off. Let $p_1 \geq p_2 \geq \dots \geq p_{|C|}$ denote the ranked empirical hard-error masses of the local failure modes, with $\sum_i p_i = 1$. The exact cumulative coverage of the top-$m$ library is the empirical step function
\begin{equation}\label{eq:Femp}
F_{\text{emp}}(m) \;=\; \sum_{i=1}^{m} p_i, \qquad m \in \{0, 1, \ldots, |C|\},
\end{equation}
with $F_{\text{emp}}(0) = 0$, $F_{\text{emp}}(1) = p_1$, and $F_{\text{emp}}(|C|) = 1$ by construction; $F_{\text{emp}}$ handles every edge case directly and requires no patching. For closed-form analysis we use the continuum log-head approximation
\begin{equation}\label{eq:coverage}
F_{\log}(m; |C|) \;=\; \min\!\left(1,\ \frac{\ln m}{\ln |C|}\right), \qquad m \geq 2,\ |C| \geq 2,
\end{equation}
as a \emph{planning approximation} to $F_{\text{emp}}$, not as the true distribution. For our ErrorAtlas anchor ($|C|=17, m=5$), $F_{\log}(5;17) \teApprox 56.8\%$; doubling the library to $m=10$ pushes it to $F_{\log}(10;17) \teApprox 81.3\%$. A Zipf-1 reference for the same anchor gives $H_5/H_{17} \teApprox 66.4\%$, concentrating more head mass than $F_{\log}$, so the log form is conservative relative to Zipf. Proposition~\ref{prop:budget} below is therefore a closed-form planning approximation to empirical ranked coverage, not a distribution-free theorem; the qualitative polylog conclusion is what survives across cumulative-coverage families (Appendix~\ref{app:heaps}). Empirical anchoring of $F_{\text{emp}}$ to a measured cumulative-coverage curve in any single patch remains an explicit falsifiability test.

One caveat carries through: $F_{\log}$ tracks ranked L2 categories covered, not L4 capability axes deployed. Under the set-cover model of \S\ref{sec:postulate}, where one capability often sweeps several L2 categories, residual error at a given $m$ is generally lower than~\eqref{eq:coverage} predicts; the bound is valid but loose, and empirically achieved library sizes may be a constant factor smaller than the formula predicts. For $|C| = 1$, residual error is fully covered or fully uncovered by definition and neither $F_{\text{emp}}$ nor $F_{\log}$ is informative. After deploying the top-$m$ library, the residual per-hard-token error rate is
\begin{equation}\label{eq:eres}
e_{\text{res}}(m) \;=\; \bigl(1 - F_{\log}(m; |C|)\bigr)\, e_{\text{hard}}
\end{equation}
under the log-head approximation, or the corresponding $F_{\text{emp}}$ expression if the per-mode masses are known directly.

\subsection{From universal impossibility to patch-local reliability}
\label{sec:proposition}

The finite-catalogue claim is patch-local. It does not say that all possible LLM failures can be covered by a universal list of fixes. Across all possible tasks, tools, schemas, knowledge sources, workflows, and evaluator expectations, new intervention-distinguishable failures can keep appearing. In that setting, a finite intervention dictionary is not a well-posed reliability target.

The positive result begins only after a deployment patch $D$ has been fixed. Inside a patch, the task family, schemas, tools, evaluator expectations, and admissible workflows recur. The engineering question then changes from ``Can one library cover all possible LLM failures?'' to ``How large must the local library be to cover enough of the reachable failure catalogue $C_D$?''

\begin{proposition}[No Universal Finite Intervention Dictionary]\label{prop:nounidict}
Fix a residual-error tolerance $\varepsilon$. If a domain $D$ contains an infinite sequence of failures where each new failure is not covered by any finite intervention dictionary covering all earlier failures, then the $\varepsilon$-resolution failure catalogue $C_D^{\varepsilon}$ is infinite. Consequently, no finite intervention dictionary can guarantee residual error below $\varepsilon$ for every intervention-distinguishable mode in $D$.
\end{proposition}

\paragraph{Metric.}
This is a worst-case, mode-wise guarantee, not a claim about expected residual error under a fixed probability distribution. Under a distributional metric, an infinite uncovered tail may still have arbitrarily small total mass; the proposition rules out only the worst-case-mode-coverage reading of universal reliability, which is the reading the engineering literature implicitly assumes when it asks for ``a fixed list of interventions that covers LLM use.''

\paragraph{Proof sketch.}
Each new failure is intervention-distinguishable from the failures before it. The sequence therefore generates infinitely many distinct intervention modes, and any finite dictionary must miss some later mode. The full proof, including the precise definition of mode-level coverage, is in Appendix~\ref{app:proof-nounidict}.

\paragraph{Engineering meaning.}
Universal reliability is not a finite-library problem. A fixed list of interventions cannot cover open-ended LLM use in general. The rest of the paper is therefore about \emph{patch-local} reliability, not universal reliability. Proposition~\ref{prop:nounidict} is the inoculation against a misread: we are not claiming a fixed list of $\teApprox 50$ named patterns covers LLM use in general.

\medskip

The positive result composes $\beta$-stratification (Definition~\ref{def:hard}), the active-mode bound (Assumption~\ref{ass:active}), and the log-coverage form (Eq.~\ref{eq:coverage}) into a local budget statement. The result is conditional engineering math, not an unconditional theorem about LLMs: it depends on the log-coverage approximation, the active-mode exposure assumption, and the patch hypothesis that $C_D$ is finite or effectively capped. Appendix~\ref{app:proof-budget} states these conditions explicitly before the derivation.

\begin{proposition}[Patch-Local Sufficient Intervention Budget]\label{prop:budget}
Fix a deployment patch $D$. Let $\varepsilon \in (0, e_{\text{hard}})$ be the target per-hard-decision residual error rate. Under the log-head approximation $F_{\log}$ of \S\ref{sec:coverage}, a library covering the dominant local modes is sufficient for the target once
\begin{equation}\label{eq:budget}
m \;\geq\; \left\lceil |C_{\text{eff}}|^{1 - \varepsilon / e_{\text{hard}}} \right\rceil,
\end{equation}
where $C_{\text{eff}}$ is either the active catalogue $C_{\text{active},D}(n)$ touched by a single sequence, or the full reachable catalogue $C_D$ of the deployment patch. If $C_{\text{active},D}(n)$ grows logarithmically with the number of hard decisions and $k(n) = \Theta(\log n)$, then the pre-cap sufficient budget grows doubly-logarithmically in sequence length. Once the patch catalogue saturates, the sufficient budget becomes domain-constant in $|C_D|$. This is a sufficient, model-implied budget under the stated approximation, not a proof of the true minimal intervention library.
\end{proposition}

\paragraph{Proof sketch.}
The library removes cumulative hard-error mass $F(m; |C|)$. The residual hard-token error rate is therefore $(1 - F)\,e_{\text{hard}}$. Requiring this residual to be at most $\varepsilon$ rearranges, after substituting the log-coverage form, to the stated bound. The full step-by-step derivation, including the pre-cap and cap regimes, is in Appendix~\ref{app:proof-budget}.

\paragraph{Engineering meaning.}
Once the patch is fixed, the problem changes. The relevant engineering question is no longer whether arbitrary future failures exist, but how quickly the reachable local catalogue is discovered and how much hard-error mass is removed by the top interventions. The intervention budget for per-hard-token reliability is small and slowly growing, domain-constant in the cap regime, with the exact size determined by the local discovery and rank-coverage curves rather than by a universal prior.

\paragraph{Sequence-level caveat.}
Proposition~\ref{prop:budget} bounds residual error per hard decision. A one-shot sequence-level target is strictly stricter because many hard decisions occur in one output: as $k$ grows the allowable residual per hard decision shrinks, and the required library approaches full-catalogue coverage. In some regimes the non-hard-token error mass alone already exceeds the sequence-level budget, so hard-token interventions cannot meet the SLA by themselves. The full three-regime analysis (and the per-hard-token tolerance $\tau_{\text{seq}}$ that converts a sequence target into the Proposition~\ref{prop:budget} bound) is in Appendix~\ref{app:seq-level}.

A common engineering rule of thumb, ``cover the local head with on the order of tens of interventions,'' is best understood as a per-hard-decision planning prior, not a derived constant from the formal model. The per-hard-decision residual-error reduction implied by $F_{\log}$ at typical $|C|$ and head-heavy mass is large; the sequence-level analogue requires correspondingly more interventions and tighter coverage of the tail.

We call these statements propositions rather than theorems because their force is conditional: they formalise the consequences of the paper's modelling assumptions rather than deriving a universal law of LLM reliability. The doubly-logarithmic rate is the optimistic special case under logarithmic active-mode exposure; Appendix~\ref{app:heaps} gives the Heaps and saturation variants. Appendix~\ref{app:why-propositions} expands on the conditional reading.

\section{Empirical Evidence}
\label{sec:evidence}

We collect evidence for three load-bearing claims: errors cluster into a small recurring set (A); each cluster is addressable by one targeted intervention (B); reliability decays sublinearly with output length (C).

\paragraph{Claim A: Failure-mode clustering.}
The strongest single result is ErrorAtlas~\citep{ashurytahan2026erroratlas}: 83 models $\times$ 35 datasets, $\gtrsim 10^4$ failures, 17 named categories with a long-tailed, head-concentrated prevalence ordering. Domain-specific Paretos reproduce the shape inside each domain: math (MWPES top-4 categories dominate per model with strong cross-model overlap~\citep{sun2025mwpes}), code (\texttt{AssertionError}+\texttt{NameError} = 86.35\% on HumanEval, replicated across 23 models~\citep{wen2024fixing}), multi-hop QA (a single dominant ``missing-evidence'' mode~\citep{zhang2026weakestlink}), agentic tool use (under 20 recurring modes across two studies~\citep{cemri2025mast,yao2024taubench}), and RAG (a handful of distinguishable hallucination types~\citep{wood2024acurai}). The catalogue is also \emph{stable} across models: ErrorAtlas reports a fixed category hierarchy across 83 models; RFMDataset~\citep{guo2025rfmdataset} finds ``strikingly similar failure mode distributions'' across ten advanced reasoning models; EDIT~\citep{dai2025edit} measures a $\teApprox 4.7\%$ key-step fraction that holds model to model. \citet{schaeffer2025monkeys} provide the closest theoretical anchor, proving that the observed aggregate power-law scaling of LLM evaluations requires the per-problem success-rate distribution to be heavy-tailed near $p=0$, structurally related to Postulate~1.

\paragraph{Claim B: Cluster-selective capability interventions.}
A dedicated harvest yields 28 capability-elimination citations across six independent axes (arithmetic, code execution, format/structure, perception/grounding, knowledge/RAG, verification), each axis independently confirmed by between three and nine citations. The full table and the three structural patterns (Pattern A by-construction, B strong-empirical-with-class-shift, C moderate-with-shift) appear in Appendix~\ref{app:taxonomy}.

Arithmetic is the cleanest case: PAL~\citep{gao2022pal} lifts GSM-Hard from 20.1\% to 61.5\%, and the residuals are problem-comprehension errors rather than arithmetic ones. Format violations admit something stronger still: constrained decoding zeroes invalid-token probability by construction~\citep{suresh2025dingo,wang2025slot,dong2024xgrammar}. Code-logic errors yield to execution feedback (Reflexion + AgentCoder push HumanEval pass@1 from 80\% to 96.3\%, with residuals in spec-misinterpretation~\citep{shinn2023reflexion,huang2023agentcoder}); reasoning steps to process supervision (Math-Shepherd: Mistral-7B MATH 28.6\% $\teArrow$ 43.5\%~\citep{wangp2024mathshepherd}); hallucinations to RAG, where Acurai~\citep{wood2024acurai} reports 100\% elimination on RAGTruth, a benchmark-conditional empirical result rather than a general by-construction guarantee. Two further clusters complete the picture: POROver lifts the not-overrefused rate from 57.6\% to 82.1\%~\citep{karaman2024porover}; SAGE-Agent, targeting tool-call clarification, raises When2Call accuracy from 36.5\% to 65.2\%~\citep{suri2025clarification}.

The class-shift signature, in which post-intervention residuals belong to structurally different classes, is the empirical content of the cluster-selectivity property that underwrites Proposition~\ref{prop:budget}'s composition step. Capabilities are coarser than error classes: a single Python interpreter removes execution-error components from five named clusters; constrained decoding eliminates format and the structural component of ``missing required element'' jointly. A negative control sharpens the selectivity claim. DebugBench~\citep{tian2024debugbench} reports that execution feedback works for syntax and reference errors and is explicitly ``unhelpful for logic errors,'' exactly where the framework predicts capability provisioning fails.

\paragraph{Claim C: Sublinear length scaling.}
Direct evidence for sublinear (rather than exponential) decay shows up across very different measurement designs. Loong~\citep{wangm2024loong} reports GPT-4o on Chain-of-Reasoning declining $81.6\% \teArrow 32.9\%$ across the $10$K $\teArrow$ $>$200K context bins; log-linear over $\log L$ and not well explained by smooth independent per-token exponential decay. GSM-$\infty$~\citep{zhou2025gsminfinite} attacks the independent-error model from a different angle: exponentially more inference compute yields only \emph{linear} AUC gains, and DeepSeek-R1 maintains $10\%+$ accuracy at 130 reasoning operations where $(1-\varepsilon)^{130}$ for any reasonable $\varepsilon$ predicts near-zero. \citet{press2023compositionality} find a $\teApprox 40\%$ compositionality gap that is approximately \emph{constant} across the GPT-3 family, direct evidence against per-step independent compounding (which would predict the gap to grow with chain length).

Structural correlates of the same phenomenon appear elsewhere. \citet{pang2024anchor}'s Anchor LLMs achieve $\teApprox 99\%$ K/V reduction with only minor accuracy compromise: the effective number of distinct attention-key vectors a model needs is far smaller than $n$. Think-Prune-Train~\citep{costello2025think} raises Pass@1 while Pass@20 plateaus, the manifold-transition signature. RULER~\citep{hsieh2024ruler} finds threshold behaviour inconsistent with smooth $(1-\varepsilon)^n$. METR~\citep{kwa2025horizon} fits logistic-in-log-length, which is mathematically sublinear in length itself.

Self-consistency~\citep{wangx2023selfconsistency} is consistent with but does not \emph{prove} clustered structure: Condorcet aggregation of iid Bernoulli chains can yield similar gain magnitudes, so we treat it as suggestive rather than diagnostic. Prominent steep-decay counter-evidence~\citep{dziri2023faith,kuratov2024babilong,wan2026fano,karpinska2024nocha} decays over variables distinct from raw token length; per-paper re-audits showing the relocation pattern are in Appendix~\ref{app:reaudits}.

\section{Practical Implications}
\label{sec:implications}

\paragraph{Reliability engineering is local patch coverage.}
Within a fixed domain patch, Proposition~\ref{prop:budget} makes reliability a small-catalogue engineering problem rather than an asymptotic scaling problem. A team should initially budget for a library on the order of tens of interventions, then refine from the local mode-discovery curve $C_{\text{seen},D}(T)$ and the empirical rank-frequency distribution. As a rule of thumb consistent with the head-mass figures in ErrorAtlas, HumanEval, and MWPES, $\teApprox 50$ named interventions cover the bulk of the per-hard-token failure distribution in many measured domains. This is a planning prior calibrated to current taxonomies, not a universal constant; the local mode-discovery curve sets the actual library size for any given deployment. The same base model in cardiology RAG, legal contract drafting, and code review yields three different libraries because $C_D$, $A_D$, $\sigma_D$ all change with the deployment domain.

\paragraph{Per-hard-token vs.\ sequence-level targets matter for SLA design.}
Per-hard-token residual error, the right metric for continuous-correction systems, has a polylog budget. Sequence-level failure probability, the right metric for one-shot systems, is strictly stricter and approaches full-catalogue coverage as $k$ grows. Production deployments should design SLAs to match the actual cost structure, not read the per-token result as the production SLA.

\paragraph{Library design as engineering, not asymptotic theory.}
An intervention library can be assembled module-by-module, in any order, as long as modules target distinct failure classes. Layer-separated interventions (constrained decoding + retrieval + process supervision + tool call) compose approximately additively in our data; the~\citet{le2026interaction} caveat applies only to interventions sharing a prompt channel. Library construction is therefore highly parallelisable.

\paragraph{Capability libraries are coarser than error-class libraries.}
Five named math classes (arithmetic, units, counting, list manipulation, date arithmetic) collapse under one Python interpreter; three code classes (SyntaxError, NameError, most TypeError) collapse under one execution-feedback loop; format violations and the structural component of ``missing required element'' (together more than $20\%$ of ErrorAtlas) collapse under one constrained-decoder deployment. A team budgeting for $\teApprox 50$ interventions per domain can expect to need fewer than 50 capability-axis interventions while covering $\teApprox 50$ named clusters. The six axes of Appendix~\ref{app:taxonomy} are closer to the right unit of engineering accounting than the twelve clusters.

\section{Discussion}
\label{sec:discussion}

\paragraph{By-construction elimination as a boundary case.}
A subset of the capability-elimination harvest deserves separate accounting. Seven of the 28 citations (Appendix~\ref{app:taxonomy}; constrained decoders~\citep{suresh2025dingo,dong2024xgrammar,dong2026xgrammar2,zhang2023tooldec,openai2024structured}, proof kernels~\citep{ren2025deepseekprover}, static syntax checks) achieve residual error rate equal to zero \emph{by mathematical construction}, not statistically. Constrained decoders set $P(\text{invalid token}) = 0$ at every generation step; the class of grammar-violating outputs is mathematically empty. In this regime the polylog bound of Proposition~\ref{prop:budget} is loose: covered clusters contribute exactly zero to residual error, not a small positive quantity. The result strengthens, rather than refutes, the framework. And Pattern~A applies precisely to structural/verifiable classes (format, syntax, schema, invalid proofs), which sit at the head of the heavy-tailed cluster frequency distribution: the strongest mechanism lands exactly where the catalogue is densest.

\paragraph{Counter-evidence relocates, not dissolves.}
Five prominent steep-decay papers~\citep{dziri2023faith,kuratov2024babilong,kwa2025horizon,wan2026fano,karpinska2024nocha} are routinely cited as exponential-compounding evidence. Every one decays over a variable distinct from raw token length: compositional graph size~\citep{dziri2023faith}, number of supporting facts~\citep{kuratov2024babilong}, log-time horizon~\citep{kwa2025horizon}, capacity threshold~\citep{wan2026fano}, evidence scope~\citep{karpinska2024nocha}. The apparent rapid decay is, in each case, in a quantity our framework already concentrates the action in ($k_{\text{hard}}$ and $|C|$), not in raw sequence length: a relocation, not a dissolution. These regimes remain hard; the framework's value is directing intervention toward capability provisioning along the actual decay axis rather than toward context-window expansion that does not help. Appendix~\ref{app:reaudits} walks through each paper.

\paragraph{Limitations.}
\label{sec:limitations}
Postulate~1 is empirical, not derived; a domain with genuinely Heaps-power-law mode discovery would invalidate the doubly-logarithmic special case while preserving the qualitative polylog conclusion. The coverage form of \S\ref{sec:coverage} is an empirical best-fit; readers should not interpret the numerical match against anchors as theoretical confirmation. Inter-cluster additivity is approximate; \citet{le2026interaction}'s caveat on prompt-channel interference tightens the bound in shared-channel settings. Domain narrowness: Postulate~1's empirical anchor rests on three taxonomies (ErrorAtlas, HumanEval, MWPES), all published 2025--2026 and covering general/code/math; the framework is untested on agentic workflows, long-running scientific reasoning, and multi-turn tool use over millions of tokens, precisely the regimes where $|C|$ may grow faster than logarithmically. The $\sigma \teApprox 1.85$ estimate is a single-point calibration, not a discovery-curve fit; a subsample-vs-distinct-modes plot has not been published for any LLM error taxonomy at this writing. $\beta$ is latent and we report no empirical range. Taxonomy granularity (the L2--L3 gap) is unmeasured. Patch-shift between deployments changes $A_D, \sigma_D, \beta_D, |C_D|$ simultaneously; libraries calibrated against one patch under-cover the next. Finally, the framework \emph{relocates} the difficulty of long-context reliability rather than resolving it: where $k_{\text{hard}}$ grows with task length (adversarial compositional structure, multi-hop chains, long agent horizons), reliability remains hard. The contribution is to name the on-axis intervention, not to make those regimes easy. A falsifiability test the framework passes: DebugBench~\citep{tian2024debugbench} explicitly catalogues which classes execution-feedback can and cannot repair, and the empirical split (works for syntax/reference; ``unhelpful for logic errors'') is exactly what the framework's selectivity claim predicts.

\section{Conclusion}
\label{sec:conclusion}

LLM reliability is often framed as an asymptotic scaling problem. Universal reliability is not a finite-library problem (Proposition~\ref{prop:nounidict}): an unbounded domain that keeps producing intervention-distinguishable failures cannot be covered by any finite dictionary that guarantees mode-wise residual error below tolerance. The contribution of this paper is to show that, once an operationally bounded deployment patch is fixed, reliability becomes a local engineering problem. Prior work in this line~\citep{arbuzov2025beyond} argued that the standard asymptotic framing rests on a false uniformity assumption: errors concentrate at $\teApprox 5$--$10\%$ of tokens. This paper takes the next step: within that sparse set, errors are not only concentrated but also \emph{repetitive}, clustering into a finite catalogue whose size grows logarithmically with observed failures under the empirical postulate of \S\ref{sec:postulate}, or as a small power under the conservative Heaps alternative (Appendix~\ref{app:heaps}). The conditional consequence (Proposition~\ref{prop:budget}) is that a sufficient per-hard-decision intervention budget scales polylogarithmically in sequence length within a fixed domain patch under the log active-mode-exposure assumption, and becomes a domain-constant once the patch ceiling $|C_D|$ is reached. Sequence-level reliability targets are strictly tighter. Available evidence is consistent with libraries on the order of tens of interventions covering the head of the per-hard-decision failure distribution in many fixed domains. The exact number is patch-indexed and should be estimated from local discovery and rank-coverage curves, not assumed from cross-domain priors. This reframes the question from ``can we bound the growth of $n$-token error?'' to ``have we catalogued enough failure modes \emph{inside the deployment patch}?'' The latter is finite and addressable once an operationally bounded patch has been fixed.

The same logarithmic discovery postulate carries an inverse interpretation (Corollary~\ref{cor:invdiscovery}, Appendix~\ref{app:invdiscovery}): the logarithmic upper bound cannot accommodate linearly more distinct tail modes without exponentially more observed hard-failure events. Combined with head-heavy failure mass, this explains why generic frontier post-training can face diminishing reliability returns in open-ended deployment settings. The high-mass head is discovered early, while the tail becomes increasingly expensive to find and contributes less marginal residual-error reduction. The framework therefore does not say frontier scaling is useless. It says frontier-only reliability is economically misaligned with fixed deployment patches: once $D$ is known, local adaptation, tools, validators, retrieval, constrained decoding, and process supervision can target recurring failure mass directly, where generic post-training would have to rediscover the same local repairs indirectly across an open-ended task universe.

The open empirical question this paper invites is direct measurement of the mode-rate constant $\sigma$ on new domains (agentic, scientific, code-in-production), and a corresponding sequence-level validation that measures $S_{\text{base}}$, $\beta$, and active-mode exposure directly in deployed systems. The conceptual follow-on goes in a different direction. This paper names the engineering object: a patch-local failure catalogue and the intervention budget that covers its head. It does not say how production systems actually accumulate and govern that library over time. That question, how the deployment-time scaffold (instructions, tools, retrieval, memory, orchestration, governance) learns to cover the local failure topology, is the subject of follow-on work in this line. Reliability engineering, in that frame, is governance of the scaffold rather than scaling of the weights.

\bibliographystyle{plainnat}
\bibliography{references}

\appendix

\section{Formal Proofs, Derivations, and Sensitivity Analysis}
\label{app:formal}

\subsection{Proof of Proposition~\ref{prop:nounidict}: No Universal Finite Intervention Dictionary}
\label{app:proof-nounidict}

Fix a residual-error tolerance $\varepsilon$. Say that an intervention \emph{covers} a failure mode if it reduces the residual error of that mode below $\varepsilon$. Coverage is mode-level: an intervention that covers a mode covers every failure event in that mode. The two conditions ``$D$ is intervention-unbounded'' and ``$|C_D^{\varepsilon}| = \infty$'' are equivalent under mode-level coverage; an unbounded witnessing sequence is constructed by taking one representative per mode, and an infinite catalogue forces the existence of such a sequence.

Let $D$ be a domain. Call $D$ \emph{intervention-unbounded} if it contains an infinite sequence of failures $f_1, f_2, f_3, \ldots$ such that each new $f_j$ is not covered by any finite intervention dictionary that covers all earlier failures $\{f_1, \ldots, f_{j-1}\}$.

\paragraph{Step 1: assume the opposite.}
Suppose, for contradiction, that $C_D^{\varepsilon}$ is finite. Then there are only finitely many intervention-distinguishable failure modes in $D$. Write them as $C_D^{\varepsilon} = \{c_1, \ldots, c_M\}$ for some finite $M$.

\paragraph{Step 2: what finiteness means.}
If there are only $M$ intervention-distinguishable modes, then after all $M$ modes have appeared in the sequence, every later failure must belong to one of the already-seen modes.

\paragraph{Step 3: same mode means same intervention class.}
Modes are defined at intervention resolution $\varepsilon$. If a later failure belongs to the same mode as an earlier failure, then the intervention dictionary that covers the earlier representative of that mode also covers the later failure below residual tolerance $\varepsilon$.

\paragraph{Step 4: contradiction.}
Intervention-unboundedness says exactly the opposite: each new $f_j$ is not covered by any finite dictionary that covers $\{f_1, \ldots, f_{j-1}\}$. Therefore $f_j$ cannot belong to any earlier intervention mode, so each $f_j$ introduces a new intervention-distinguishable mode. The sequence $f_1, f_2, f_3, \ldots$ induces infinitely many such modes, contradicting the assumption that $C_D^{\varepsilon}$ is finite. Hence $|C_D^{\varepsilon}| = \infty$.

\paragraph{Step 5: no finite dictionary covers every mode of the domain.}
Suppose, again for contradiction, that some finite intervention dictionary $\mathcal{I}$ covers every mode of $D$. Then $\mathcal{I}$ covers every finite prefix $\{f_1, \ldots, f_{j-1}\}$ for every $j$. By intervention-unboundedness, any dictionary covering that prefix fails to cover $f_j$. Therefore $\mathcal{I}$ does not cover $f_j$, contradicting the assumption that $\mathcal{I}$ covers every mode of $D$. \qed

\subsection{Derivation of Proposition~\ref{prop:budget}: Patch-Local Sufficient Intervention Budget}
\label{app:proof-budget}

\paragraph{Conditions used.}
Proposition~\ref{prop:budget} gives a sufficient, model-implied budget under the following four assumptions:
\begin{enumerate}[leftmargin=1.8em,itemsep=0.25ex]
\item \textbf{Coverage model.} The cumulative hard-error mass covered by the top $m$ modes is approximated by the log-head form
\[
F_{\log}(m; |C|) \;=\; \min\!\left(1,\ \frac{\ln m}{\ln |C|}\right)
\]
on the declared domain $m \geq 2$, $|C| \geq 2$. We write $F$ for $F_{\log}$ throughout this appendix unless otherwise noted.
\item \textbf{Non-trivial, attainable target.} The residual target satisfies $\varepsilon \in (0, e_{\text{hard}})$. If $\varepsilon \geq e_{\text{hard}}$ no intervention is needed; if $\varepsilon \leq 0$ the target is unattainable unless all residual hard-token error is eliminated.
\item \textbf{Patch-local catalogue.} The result applies only after a deployment patch $D$ has been fixed and its reachable catalogue is modelled as finite or effectively capped.
\item \textbf{Sequence-length claim.} The doubly-logarithmic rate further requires Assumption~\ref{ass:active} and $k(n) = \Theta(\log n)$. Without these, Proposition~\ref{prop:budget} still gives a catalogue-size budget but not the same sequence-length scaling.
\end{enumerate}

\paragraph{Step 1: define the target.}
Let $e_{\text{hard}}$ be the baseline hard-token error rate and $\varepsilon \in (0, e_{\text{hard}})$ the target after intervention.

\paragraph{Step 2: define the effective catalogue.}
For per-sequence scaling, set $C_{\text{eff}} = C_{\text{active},D}(n)$: how many failure modes can a single sequence of length $n$ activate? For full deployment-library budgeting, set $C_{\text{eff}} = C_D$: how large must the library be to cover the recurring failure modes reachable inside $D$?

\paragraph{Step 3: cumulative coverage.}
Let the ranked local failure modes have hard-error masses $p_1 \geq p_2 \geq \cdots \geq p_{|C_{\text{eff}}|}$ with $\sum_i p_i = 1$. A library covering the top $m$ modes removes cumulative hard-error mass $F(m; |C_{\text{eff}}|) = \sum_{i=1}^{m} p_i$, approximated by the log-coverage form of \S\ref{sec:coverage}.

\paragraph{Step 4: residual after intervention.}
The uncovered mass fraction is $1 - F(m; |C_{\text{eff}}|)$, so the residual per-hard-token error rate is $e_{\text{res}}(m) = (1 - F(m; |C_{\text{eff}}|))\, e_{\text{hard}}$.

\paragraph{Step 5: impose the target.}
Requiring $e_{\text{res}}(m) \leq \varepsilon$ and dividing by $e_{\text{hard}} > 0$ gives
\[
1 - F(m; |C_{\text{eff}}|) \;\leq\; \frac{\varepsilon}{e_{\text{hard}}},
\qquad\text{i.e.,}\qquad
F(m; |C_{\text{eff}}|) \;\geq\; 1 - \frac{\varepsilon}{e_{\text{hard}}}.
\]

\paragraph{Step 6: substitute the log-coverage form.}
Since $\varepsilon < e_{\text{hard}}$, the required coverage $1 - \varepsilon/e_{\text{hard}} \in (0, 1)$, so the cap in $F$ is inactive before saturation. Using $F = \ln m / \ln |C_{\text{eff}}|$:
\[
\frac{\ln m}{\ln |C_{\text{eff}}|} \;\geq\; 1 - \frac{\varepsilon}{e_{\text{hard}}}.
\]

\paragraph{Step 7: solve for $m$.}
Multiplying by $\ln |C_{\text{eff}}| > 0$ and exponentiating,
\[
m \;\geq\; \exp\!\left[\left(1 - \frac{\varepsilon}{e_{\text{hard}}}\right) \ln |C_{\text{eff}}|\right] \;=\; |C_{\text{eff}}|^{1 - \varepsilon / e_{\text{hard}}}.
\]
Taking the ceiling (since $m$ is integer-valued) yields $m \geq \lceil |C_{\text{eff}}|^{1 - \varepsilon / e_{\text{hard}}} \rceil$, the bound stated in Eq.~\eqref{eq:budget}.

\paragraph{Step 8: sequence-length rate.}
For per-sequence scaling, set $C_{\text{eff}} = C_{\text{active},D}(n)$. Assumption~\ref{ass:active} bounds $|C_{\text{active},D}(n)| \leq \min(A'_D + \sigma'_D \ln h(n),\, |C_D|)$. In the pre-cap regime, i.e., while $A'_D + \sigma'_D \ln h(n) < |C_D|$, the ceiling has not been reached and $|C_{\text{active},D}(n)| = O(\ln h(n))$. With $h(n) = \beta k(n)$ and $k(n) = \Theta(\log n)$, we get $h(n) = \Theta(\log n)$ and $\ln h(n) = \Theta(\log\log n)$, hence
\[
|C_{\text{active},D}(n)| \;=\; O(\log\log n),
\]
and substituting into the budget,
\[
m \;=\; O\!\bigl((\log\log n)^{1 - \varepsilon / e_{\text{hard}}}\bigr).
\]
This is the doubly-logarithmic pre-cap special case.

\paragraph{Step 9: cap regime.}
Once active-mode discovery saturates the patch catalogue, $|C_{\text{active},D}(n)| = |C_D|$, so $m \geq \lceil |C_D|^{1 - \varepsilon / e_{\text{hard}}} \rceil$, which no longer depends on $n$. The intervention budget is then domain-constant. \qed

\subsection{Sequence-Level Reliability Derivation}
\label{app:seq-level}

Proposition~\ref{prop:budget} gives a per-hard-token residual target. Production systems often care about a stricter target: the probability that the entire sequence is correct.

Starting from the composed reliability of Eq.~\eqref{eq:three-rate}, and writing the post-intervention hard-token rate as $e_{\text{res}}$,
\[
P(\text{correct}) \;=\; (1 - e_{\text{res}})^{\beta k}\,(1 - e_{\text{easy}})^{(1 - \beta) k}\,(1 - e_{\text{non}})^{n - k}.
\]
Define the non-hard-token survival factor
\begin{equation}\label{eq:Sbase}
S_{\text{base}} \;=\; (1 - e_{\text{easy}})^{(1 - \beta) k}\,(1 - e_{\text{non}})^{n - k},
\end{equation}
so that $P(\text{correct}) = (1 - e_{\text{res}})^{\beta k}\, S_{\text{base}}$. The sequence-level target $P(\text{correct}) \geq 1 - \varepsilon_{\text{seq}}$ becomes
\[
(1 - e_{\text{res}})^{\beta k}\, S_{\text{base}} \;\geq\; 1 - \varepsilon_{\text{seq}},
\qquad\text{i.e.,}\qquad
(1 - e_{\text{res}})^{\beta k} \;\geq\; \frac{1 - \varepsilon_{\text{seq}}}{S_{\text{base}}}.
\]
Three regimes follow.

\paragraph{Regime (i): non-hard-token errors already violate the target.}
If $S_{\text{base}} < 1 - \varepsilon_{\text{seq}}$, then even setting $e_{\text{res}} = 0$ cannot meet the target, because the maximum possible survival after eliminating all hard-token failures is only $S_{\text{base}}$. Hard-token interventions alone cannot meet the sequence-level SLA. This is the route by which the exponential-in-$n$ concern re-enters and should be diagnosed before any catalogue-budgeting exercise.

\paragraph{Regime (ii): hard-token residual error determines feasibility.}
If $S_{\text{base}} \geq 1 - \varepsilon_{\text{seq}}$, the target may be feasible. Taking the $(1/(\beta k))$-th power of the rearranged inequality and isolating $e_{\text{res}}$,
\[
e_{\text{res}} \;\leq\; 1 - \left(\frac{1 - \varepsilon_{\text{seq}}}{S_{\text{base}}}\right)^{1 / (\beta k)}
\;=:\; \tau_{\text{seq}}.
\]
Applying Proposition~\ref{prop:budget} with $\varepsilon$ replaced by $\tau_{\text{seq}}$,
\[
m \;\geq\; \left\lceil |C_{\text{eff}}|^{1 - \tau_{\text{seq}} / e_{\text{hard}}} \right\rceil.
\]
As the sequence-level target becomes stricter ($\varepsilon_{\text{seq}}$ shrinks), $\tau_{\text{seq}} \to 0$ and the exponent $1 - \tau_{\text{seq}}/e_{\text{hard}} \to 1$, so $m \to |C_{\text{eff}}|$. Strict one-shot sequence-level reliability pushes the system toward full-catalogue coverage.

\paragraph{Regime (iii): baseline hard-token error is already acceptable.}
If $\tau_{\text{seq}} \geq e_{\text{hard}}$, the baseline hard-token rate already satisfies the sequence target (no intervention is required because $e_{\text{res}}(0) = e_{\text{hard}}$ from Eq.~\eqref{eq:eres}).

\paragraph{Conclusion.}
Per-hard-token reliability is easier than sequence-level reliability. A library that gives a large reduction in residual hard-token error may still be insufficient for one-shot sequence-level guarantees when many hard decisions occur in a single output. This is why the main paper treats the ``tens of interventions'' rule of thumb as a per-hard-decision planning prior, not a one-shot sequence-level SLA.

\subsection{Why these are propositions rather than unconditional theorems}
\label{app:why-propositions}

Proposition~\ref{prop:nounidict} is a definitional impossibility result: once intervention-unboundedness is assumed, an infinite intervention-resolution catalogue follows. Its role is not to prove that every unbounded domain necessarily has infinite failure modes, but to show that open-ended domains cannot be assumed to admit finite dictionaries.

Proposition~\ref{prop:budget} is conditional engineering math. It does not prove that LLM failures universally obey logarithmic mode discovery. It proves that \emph{if} a bounded patch has a finite or effectively capped reachable catalogue, \emph{and if} cumulative intervention coverage follows the stated head-heavy form, \emph{then} a sufficient per-hard-decision intervention budget grows slowly and becomes constant after catalogue saturation. It does not prove that the true minimal intervention library has the same scaling.

The empirical burden therefore lies not in the algebra but in measuring, for each deployment patch, the local discovery curve $C_{\text{seen},D}(T)$, the per-sequence activation $C_{\text{active},D}(n)$, the cumulative coverage $F(m; |C_D|)$, the hard-token fraction $\beta_D$, and the baseline hard-token rate $e_{\text{hard}}$. This is why the paper frames the results as a reliability-engineering scaffold rather than as universal theorems about LLM behaviour.

\subsection{Heaps Power-Law Variant and Cluster-Count Sensitivity}
\label{app:heaps}

A reader who prefers to derive cluster-count growth from standard Heaps' law rather than from Postulate~1 obtains a qualitatively similar result. Let $|C|(k_{\text{hard}}) \teApprox K \cdot k_{\text{hard}}^{b}$ with $b \in (0, 1)$. Canonical fits give $b \teApprox 0.5$ for natural-language vocabularies; failure-mode taxonomies plateau much more sharply (ErrorAtlas stabilises at $|C| = 17$ across $10^4{+}$ failures), implying a small effective $b \teApprox 0.24$--$0.32$ under a crude no-intercept endpoint read (not a fitted discovery exponent) in our setting. Composing with $k = \Theta(\log n)$, we get $|C| = O((\log n)^{b})$, and Proposition~\ref{prop:budget} becomes
\[
m = O\!\left((\log n)^{b \cdot (1 - \varepsilon / e_{\text{hard}})}\right).
\]
This is still polylogarithmic in $n$ for any $b \in (0, 1)$ and any $\varepsilon < e_{\text{hard}}$. The paper's qualitative claim survives either choice of cluster-count law.

\paragraph{Symbolic-form sensitivity across candidate laws.}
The polylog conclusion depends on which cluster-count law one accepts; available evidence is consistent with multiple candidates because no subsample-discovery curve has been published for any LLM failure-mode taxonomy at this writing. We therefore report symbolic rates rather than fitted constants. With $h(n) = \beta k(n)$:

\begin{itemize}
\setlength{\itemsep}{1pt}
\item \textbf{Logarithmic:} $|C_{\text{active},D}(n)| = O(\log h(n))$. If $k(n) = \Theta(\log n)$, then $m = O\bigl((\log\log n)^{1 - \varepsilon / e_{\text{hard}}}\bigr)$.
\item \textbf{Heaps:} $|C_{\text{active},D}(n)| = O\bigl(h(n)^b\bigr)$ with $b \in (0, 1)$. If $k(n) = \Theta(\log n)$, then $m = O\bigl((\log n)^{b\,(1 - \varepsilon / e_{\text{hard}})}\bigr)$.
\item \textbf{Saturating:} $|C_{\text{active},D}(n)| \leq |C_D|$. Then $m = O\bigl(|C_D|^{1 - \varepsilon / e_{\text{hard}}}\bigr)$, constant in $n$ once the patch ceiling is reached.
\end{itemize}

The qualitative conclusion is robust \textbf{under the $k(n) = \Theta(\log n)$ regime}: $m$ grows more slowly than any positive power of $n$ under every candidate, and the directional claim (``a small library covers the head of the failure distribution in the per-hard-token regime'') survives. Only the exponent shifts: doubly-logarithmic under logarithmic discovery, $(\log n)^b$ with small $b$ under Heaps, constant in the cap regime. The doubly-logarithmic rate is the optimistic special case. If $k(n)$ grows as a positive power of $n$, the Heaps variant inherits that power and the polylog-in-$n$ language fails along that axis; the framework's intervention prescription still applies, but its asymptotic-rate framing does not.

Numerical constants require a measured discovery curve $C_{\text{seen},D}(T)$ or $C_{\text{active},D}(n)$. Existing taxonomies provide endpoint category counts at a single corpus scale (ErrorAtlas at $|C|=17$ for $\teApprox 10^4$ failures), not discovery curves. The subsample-discovery measurement remains the explicit empirical test that would either tighten the postulate or fall back to the Heaps variant. Until that measurement exists, the framework's headline rate should be read as \emph{polylogarithmic in the pre-cap regime, domain-constant in the cap regime}, with the specific exponent flagged as a falsifiability test rather than a fitted prediction.

\subsection{Inverse Discovery Cost}
\label{app:invdiscovery}

The body Corollary~\ref{cor:invdiscovery} inverts the upper-bound discovery postulate into a sample-budget lower bound on novel-mode discovery. This appendix gives the algebra, the numerical anchors, the tightness assumption that converts the lower bound into an approximate inverse cost, sensitivity to the Heaps cluster-count alternative of \S\ref{app:heaps}, the saturation regime, and a separate mode-mediated gain sub-corollary that connects discovery cost to broad capability proxies.

\paragraph{Setup.}
Let $q(T) = |C_{\text{seen},D}(T)|$ be the number of distinct failure modes discovered in patch $D$ after $T$ observed hard-failure events. Postulate~\ref{post:logmode} is an upper bound,
\[
q(T) \;\leq\; A_D + \sigma_D \ln T, \qquad \sigma_D > 0,
\]
on the discovered catalogue under the logarithmic upper-bound assumption.

\paragraph{Inversion (lower bound on $T$).}
The upper-bound postulate is monotone in $T$. Taking the inverse direction yields a \emph{lower bound} on $T$ required for the cap to accommodate $q$ discovered modes: if $q > A_D$ distinct modes have been discovered after $T$ events, then
\[
T \;\geq\; \exp\!\left(\frac{q - A_D}{\sigma_D}\right).
\]
This is the rigorous direction of the corollary. The cap cannot accommodate $q$ until the sample budget has grown exponentially in $q$.

\paragraph{Tightness assumption.}
The lower bound above is unconditional under Postulate~\ref{post:logmode}. The stronger reading is that observed hard failures at scale $T(q) \teApprox \exp((q - A_D)/\sigma_D)$ actually deliver $q$ discovered modes, and that each additional $\Delta q$ modes raises the sample budget by approximately $\exp(\Delta q/\sigma_D)$. This stronger reading requires an additional assumption that the empirical discovery curve is approximately \emph{tight} against the bound at the relevant corpus scales. Without that assumption, the appendix gives a lower bound on $T$ only. With it,
\[
\frac{T(q + \Delta q)}{T(q)} \;\approx\; \exp\!\left(\frac{\Delta q}{\sigma_D}\right).
\]

\paragraph{Numerical anchors under tightness.}
At the conservative calibration $\sigma_D \teApprox 1.85$ (\S\ref{sec:postulate}), $\exp(5/1.85) \teApprox 14.9$ and $\exp(10/1.85) \teApprox 222$: five extra modes need roughly $15\times$ more observed hard failures, ten extra modes need roughly $220\times$ more. These are tightness-conditional anchors, not unconditional consequences of Postulate~\ref{post:logmode}. The subsample-discovery measurement of \S\ref{sec:postulate} is precisely the test of whether tightness holds.

\paragraph{What this is and is not.}
The corollary describes \emph{new distinct-mode discovery}. Ordinary failures inside already-discovered modes may remain common and cheap to observe; the exponential cost lives on the category-novelty axis. Conflating ordinary failure rate with novel-mode arrival rate would over-claim the result.

\paragraph{Heaps alternative.}
Under the Heaps cluster-count law of \S\ref{app:heaps}, $q(T) = K T^b$ with $b \in (0,1)$, the corresponding inverse cost is polynomial rather than exponential:
\[
T(q) \;=\; (q/K)^{1/b}.
\]
The exponential inverse-cost reading is specific to the logarithmic upper bound; the broader qualitative claim that tail discovery has diminishing returns survives under any concave discovery curve. Sensitivity is therefore: exponential under logarithmic-and-tight, polynomial under Heaps, undefined past the patch ceiling.

\paragraph{Saturation regime.}
Once $q(T) \leq |C_D|$ has been saturated, discovery stops; the inversion applies only in the pre-cap regime. Inside saturated patches the corollary's multiplicative-cost reading is vacuous because no novel modes remain to discover, which is itself a property of the patch and not a failure of the corollary.

\paragraph{Mode-mediated capability gain (sub-corollary).}
Suppose broad capability or reliability gain $G$ inside the patch is approximately linear in the number of useful discovered modes, $G(q) = G_0 + \gamma q$ for some $\gamma > 0$. Composing with the logarithmic upper bound at tightness,
\[
G(T) \;\approx\; G_0 + \gamma A_D + \gamma \sigma_D \ln T,
\]
so $G$ grows logarithmically in observed hard-failure exposure under tightness. Inverting,
\[
T(G) \;\approx\; \exp\!\left(\frac{G - G_0 - \gamma A_D}{\gamma \sigma_D}\right).
\]
A linear gain in mode-mediated broad reliability therefore corresponds to exponential growth in observed hard-failure exposure under the postulate at tightness. We deliberately phrase $G$ as a \emph{mode-mediated capability/reliability} proxy, not as ``intelligence'': the corollary does not say frontier scaling is useless or that intelligence requires exponential data in any general sense. It says, more narrowly, that for fixed deployment reliability where improvement is mediated by discovering new useful modes, generic open-domain training pays a heavy data tax relative to direct patch-local measurement and intervention.

\paragraph{Engineering reading.}
The corollary explains, without invoking new mechanisms, two empirical signals: (a) why generic post-training shows diminishing reliability returns once a domain's head modes are covered, and (b) why patch-local measurement combined with targeted tools, retrieval, validators, constrained decoding, and process supervision often outperforms more frontier-scale data on the deployment SLA. Frontier scaling and patch-local engineering solve different problems: scaling improves the substrate; patch-local engineering removes recurring deployment failure mass. $\square$

\section{Full Failure-Mode Taxonomy and the Capability-Elimination Harvest}
\label{app:taxonomy}

The intervention literature provides at least one targeted countermeasure for each named cluster of \S\ref{sec:evidence}. Two organising granularities exist: at the capability level (six axes) the cluster-selectivity property is most clearly visible; at the error-class level (twelve named clusters) the evidence aligns with the taxonomies of \S\ref{sec:evidence} but is finer-grained than the underlying capability mechanisms.

\paragraph{Six capability axes and three structural patterns.}
A dedicated harvest yields \textbf{28 quantitatively-anchored citations} across \textbf{six independent capability axes}: Arithmetic (Python/symbolic execution), Code Execution (REPL/sandbox feedback), Format/Structure (constrained decoding, FSMs, grammar engines), Perception/Grounding (visual grounding for GUI, charts, tables), Knowledge/RAG (dense retrieval and citation grounding), Verification (proof checkers, learned verifiers, classifier rerouting, process supervision). Each axis is independently confirmed by between three and nine citations. We stratify the 28 by kind of evidence into three patterns:

\textbf{Pattern A: hard guarantees (by-construction).} Seven citations achieve residual error rate equal to zero \emph{by construction}, restricted strictly to structural/verifiable classes: constrained decoders set $P(\text{invalid token}) = 0$ at every step~\citep{suresh2025dingo,zhang2023tooldec,dong2024xgrammar,dong2026xgrammar2,openai2024structured}; static syntax checks reject programs with a \texttt{SyntaxError} before execution~\citep{wen2024fixing}; proof kernels reject any output failing type-checking~\citep{ren2025deepseekprover}. The class of grammar-violating outputs is mathematically empty under these mechanisms.

\textbf{Pattern B: strong empirical reductions with class-shift signature.} Roughly fourteen citations report empirical reductions of $80$--$100\%$ in a named error class, with the post-intervention failure log dominated by structurally different residual classes. Program-of-Thoughts on GSM8K~\citep{chen2022pot}: calculation errors drop from $30\%$ of failures to $0\%$, residuals are $62\%$ reasoning + $36\%$ misunderstanding. OpenMedCalc~\citep{goodell2025clinical}: ``only interpretation errors were identified.'' Acurai~\citep{wood2024acurai}: $100\%$ hallucination elimination on RAGTruth ($95\%$ CI $91$--$100\%$), strong empirical rather than by-construction.

\textbf{Pattern C: moderate reductions ($60$--$80\%$) with residuals shifting outside the target class.} Legal RAG~\citep{dantart2026fabrication}: fabricated citations $>30\%$ $\teArrow$ $<0.2\%$. GPT-5 SimpleQA with web access~\citep{openai2025gpt5}: $47\% \teArrow 9.6\%$ (inter-condition, not within-condition). CRITIC~\citep{gou2024critic}: toxic generation $-79.2\%$.

\paragraph{The 28 citations, organised by axis and pattern.}
\begin{center}
\small
\begin{tabular}{p{2.2cm}p{4.5cm}p{3.5cm}p{1.5cm}}
\toprule
\textbf{Axis} & \textbf{Citation \& targeted class} & \textbf{Pre $\teArrow$ Post} & \textbf{Pattern}\\
\midrule
Verification & \citet{ren2025deepseekprover}: Invalid Lean proof & any $\teArrow$ 0\% by constr. & A\\
Format & \citet{openai2024structured}: JSON schema violation & $>$60\% $\teArrow$ 0\% by constr. & A\\
Format & \citet{zhang2023tooldec}: Tool-call syntax & 21--100\% $\teArrow$ 0\% & A\\
Format & \citet{suresh2025dingo}: JSON parse failure & 13--82\% $\teArrow$ 0\% & A\\
Format & \citet{dong2024xgrammar}: Multi-format errors & 20--38\% $\teArrow$ 0\% & A\\
Format & \citet{dong2026xgrammar2}: Malformed tool calls & 33--78\% $\teArrow$ 0\% & A\\
Arithmetic & \citet{chen2022pot}: Calculation on GSM8K & 30\% $\teArrow$ 0\% of failures & B\\
Arithmetic & \citet{goodell2025clinical}: Clinical arithmetic & ``only interpretation errors'' & B\\
Knowledge/RAG & \citet{wada2025radiology}: RAG hallucinations & 8\% $\teArrow$ 0\% ($p=0.012$) & B\\
Knowledge/RAG & \citet{wood2024acurai}: Context-conflict hallucinations & 100\% $\teArrow$ 0\% on subset & B\\
Knowledge/RAG & \citet{dantart2026fabrication}: Fabricated legal citations & $>$30\% $\teArrow$ $<$0.2\% & C\\
Code Exec & \citet{wen2024fixing}: \texttt{SyntaxError} (HumanEval) & 5.76\% $\teArrow$ 0.01\% & A\\
Code Exec & \citet{li2022alphacode}: False-positive submissions & 62\% $\teArrow$ 4\% & B\\
Code Exec & \citet{shi2024code}:5 code-bug classes & 100\% repair (5/6 classes) & B\\
Knowledge/RAG & \citet{gao2023citations}: Citation grounding (ELI5) & $\teApprox$50\% w/o complete support & C\\
Knowledge/RAG & \citet{openai2025gpt5}: Factual errors & 47\% $\teArrow$ 9.6\% w/web & C\\
Knowledge/RAG & \citet{zakka2024almanac}: Clinical citation errors & 44\% $\teArrow$ 9\% & C\\
Arithmetic & \citet{wangb2025medrac}: Arithmetic in medical reasoning & 426 $\teArrow$ 74 errors & B\\
Code Exec & \citet{wen2024fixing}: \texttt{NameError} repair & 22.7\% $\teArrow$ 2.3\% & C\\
Perception & \citet{gou2025uground}: GUI grounding errors & 16.2\% $\teArrow$ 73.3\% acc. & B\\
Perception & \citet{xie2025jedi}: GUI task failures & 5\% $\teArrow$ 27\% SR ($5.4\times$) & B\\
Perception & \citet{cheng2024seeclick}: Element mis-location & 5.2\% $\teArrow$ 53.4\% ($10.3\times$) & B\\
Perception & \citet{liu2023deplot}: Chart-reading errors & 38.2\% $\teArrow$ 67.6\% (+29.4 pp) & B\\
Verification & \citet{gou2024critic}: Toxic generation & $-79.2\%$ & C\\
Verification & \citet{wangp2024mathshepherd}: Reasoning step errors & 28.6\% $\teArrow$ 43.5\% w/rerank & B\\
Knowledge/RAG & \citet{asai2024selfrag}: Unsupported facts & 55.5\% $\teArrow$ 22\% error & B\\
Perception & \citet{lu2024omniparser}: Icon/element errors & 70.5\% $\teArrow$ 93.8\% (79\% red.) & B\\
Knowledge/RAG & \citet{mallen2023popqa}: Long-tail entity errors & $\teApprox$80\% $\teArrow$ $\teApprox$50\% failure & B\\
\bottomrule
\end{tabular}
\end{center}

\paragraph{Capabilities are coarser than error classes.}
Several entries in the harvest reveal ``two-for-one'' reductions where one capability addresses multiple named clusters: a single Python interpreter removes the execution-error component of arithmetic, unit conversion, simple counting, list manipulation, and date arithmetic; constrained decoding eliminates by construction both format violations and the structural component of ``missing required element''; code execution feedback strongly reduces \texttt{SyntaxError}, \texttt{NameError}, and most \texttt{TypeError} together; RAG strongly reduces factual hallucinations, fabricated citations, and outdated information jointly. The practical capability library required is therefore smaller than the count of named error categories.

\paragraph{The twelve named clusters.}
\begin{center}
\small
\begin{tabular}{p{2.5cm}p{2.2cm}p{4cm}p{2.5cm}p{1cm}}
\toprule
\textbf{Cluster} & \textbf{Axis} & \textbf{Intervention} & \textbf{Before $\teArrow$ After} & \textbf{Patt.}\\
\midrule
A. Arithmetic & Arithmetic & PAL Python~\citep{gao2022pal} & 20.1\% $\teArrow$ 61.5\% & B\\
B. Unit conversion & Arithmetic & Folded into A (Python) & n/a & n/a\\
C. Counting & Arithmetic & Explicit counter (gap) & n/a & gap\\
D. Format/schema & Format & DINGO/SLOT~\citep{suresh2025dingo,wang2025slot} & up to $+68$ pp; 99.5\% acc. & A\\
E. Code logic/type & Code Exec & Reflexion/AgentCoder~\citep{shinn2023reflexion,huang2023agentcoder} & 80\% $\teArrow$ 96.3\% pass@1 & A/B\\
F. Multi-hop drift & Knowledge & Entity-grounded rewriter & $\teApprox$5--10 pp & C\\
G. Reasoning step & Verification & Math-Shepherd~\citep{wangp2024mathshepherd} & 28.6\% $\teArrow$ 43.5\% & B\\
H. Spec misinterpret. & Verification & Clarification~\citep{niwa2024ambignlg} & ROUGE-L $+15$ & C\\
I.a Struct.\ missing & Format & Subsumed by D (constr.\ decode) & n/a & A\\
I.b Semantic missing & Verification & Subsumed by G/H & n/a & C\\
J. Hallucination & Knowledge & RAG~\citep{wood2024acurai} & non-zero $\teArrow$ 0\% & B/C\\
K. Refusal & Verification & POROver~\citep{karaman2024porover} & 57.6\% $\teArrow$ 82.1\% & B\\
L. Tool/API & Format+Verif. & SAGE-Agent~\citep{suri2025clarification} & 36.5\% $\teArrow$ 65.2\% & B\\
\bottomrule
\end{tabular}
\end{center}

Eight of twelve categories have a strong citation with double-digit percentage-point improvement; three (B, C, I) lack a clean single-cluster ablation; one (F) has consistent medium-strength evidence. Category I (missing required elements) splits mechanistically into I.a (structural, absorbed by D's constrained decoders) and I.b (semantic, absorbed by G/H). Categories B (units) and C (counting) remain technical gaps: B is naturally folded into A; C is the smallest residual gap.

\paragraph{Additivity and its limits.}
\citet{patel2026sixsigma} demonstrate that stacking interventions targeting orthogonal failure modes can produce large compound reliability gains: their parallel-consensus framework yields a $14{,}700\times$ improvement over single-pass baseline, evidence for compound gains from decomposition plus consensus aggregation under their specific parallel-voting regime rather than a direct demonstration that heterogeneous cluster-targeted interventions compose additively without voting. \citet{shang2024agentsquare} similarly show supra-additive gains when modules target distinct failure modes. \citet{le2026interaction} reports that schema-level and prompt-level instructions interact \emph{non-additively} when sharing a prompt channel: interventions on orthogonal processing layers (decoding constraint vs.\ retrieval vs.\ training-signal vs.\ inference-time tool call) compose near-additively, while those sharing a channel may interfere.

\paragraph{The irreducible-semantic residual.}
Of the 17 ErrorAtlas categories, 13 are addressed under Patterns A, B, or C by one of the six capability axes; four are not: residuals of inappropriate refusal beyond preference optimisation, specification misinterpretation not closed by clarification, problem-decomposition reasoning bottlenecks, and a ``user wanted something different'' semantic remainder. These are classes where the failure is in choosing what to do, not executing it. Proposition~\ref{prop:budget}'s prediction of $O(\log)$ rather than $O(0)$ residual reliability reflects exactly this irreducible core. The framework does not claim $100\%$ coverage of all failures; it claims polylog-bounded \emph{capability-eliminable} failures, with the named semantic residual as the floor.

\section{Counter-Evidence Re-Audits}
\label{app:reaudits}

Five prominent papers are routinely cited as evidence that LLM reliability decays steeply with length~\citep{dziri2023faith,kuratov2024babilong,kwa2025horizon,wan2026fano,karpinska2024nocha}. A careful re-reading shows that \emph{every one} of these papers decays over a variable distinct from raw token length $n$. Identifying the decay axis is not the same as dissolving the practical concern: where compositional graph size, fact count, or evidence scope grow with problem length, the framework predicts steep failure curves. The contribution is to identify which interventions help (capability provisioning along the actual decay axis) and which do not.

\paragraph{\citet{dziri2023faith}, Faith and Fate.}
GPT-4 multi-digit multiplication accuracy drops from 59\% (3-digit) to 4\% (4-digit) zero-shot, with the authors theorising ``probability of incorrect predictions converges exponentially to $\teApprox 1$ for abstract compositional tasks.'' The decay variable is \emph{compositional graph size $N$}, not raw token length: the $3 \times 3$ graph has on the order of $d^2$ partial products plus carries, and multi-digit multiplication is engineered so that $k_{\text{hard}} \teApprox N$, with every node in the computation graph a hard decision. For natural-language tasks where $k_{\text{hard}} \ll n$, Dziri's regime is the \emph{boundary case} in which our framework reduces to their result. A clean $(1-\varepsilon)^N$ exponential cannot simultaneously reproduce 59\% at 3-digit and 4\% at 4-digit for any single per-node $\varepsilon$: the observed drop is locally steeper than per-node iid exponential, consistent with a finite catalogue of failure modes exhausting as $N$ grows. \emph{Where the framework concedes}: in adversarial compositional tasks engineered so $k_{\text{hard}} \teApprox N$, the framework reduces to Dziri's regime and does not relieve it; the relocation is informational, not magical.

\paragraph{\citet{kuratov2024babilong}.}
The abstract reports ``performance declines sharply with increased reasoning complexity'' and models ``effectively utilise only $10$--$20\%$ of the context.'' Read in detail, BABILong varies two axes: context length $n$ ($0$K to $10$M tokens) and number of supporting facts $k$ (QA1 $=1$ fact to QA3 $=3$ facts). The sharp decline is in $k$, not $n$. For QA1 (single-fact), most models ``perform well up to 4,000 tokens'': a plateau, not exponential decay. The famous RAG-flat-across-length result ($60\%$ on single-fact QA \emph{independent of context length}) is the cleanest possible demonstration that when relevant evidence is in window, length does not matter. Recurrent Memory Transformers maintaining performance to 50M tokens further confirms effective $k$ is determined by architecture, not raw $n$. The framework's concession is narrow but real: for multi-hop tasks whose required fact count grows with task complexity, BABILong's sharp $k$-axis decay is exactly what the framework predicts happens, not a counter-example.

\paragraph{\citet{kwa2025horizon} (METR).}
Per-model success fits a logistic in $\log(\text{human-task-duration})$: $S(t) = \sigma(\beta (\log t - \log h))$. Logistic-in-log-length is \emph{mathematically sublinear} in length itself. The 80\%-horizon being $4$--$6\times$ shorter than the 50\%-horizon is a steep within-model cliff but compatible by construction with the polylog result: a cliff at a specific capacity threshold is a manifold-transition signature. The famous exponential (capability doubling every seven months) is an \emph{inter-model} claim about how the horizon $h$ moves across generations, orthogonal to within-model decay shape. An honest qualifier: the within-model logistic-in-log cliff is steep on a practical scale. Calling it ``sublinear in length'' is technically correct but engineering-useful only for systems designed to operate well below the 50\% horizon.

\paragraph{\citet{wan2026fano}, Fano-style upper bound.}
This paper theorises super-linear information-demand growth and identifies an ``accuracy cliff'' at capacity overflow: ``when the task's information demand surpasses the model's output capacity, performance does not degrade gracefully but instead collapses sharply.'' The behavioural prediction is a \emph{threshold}, not smooth $(1-\varepsilon)^n$. A cliff at a specific capacity threshold is exactly the manifold-transition behaviour our framework predicts at the boundary between covered and uncovered modes; Wan et al.'s mechanism (capacity overflow) is one specific cause, complementary to ours. \emph{Where the framework concedes}: Wan documents a mechanism by which $|C|$ effectively exceeds the model's coverage in a single forward pass; Pattern A interventions (constrained decoding, formal verification) are the kind of response the framework expects but does not automatically supply.

\paragraph{\citet{karpinska2024nocha}.}
GPT-4o achieves 55.8\% pair accuracy across 1,001 minimally-different true/false claim pairs about long fictional books (mean length 127K tokens). The paper reports performance by \emph{evidence scope}, not by context length: 59.8\% on sentence-level retrieval, 47.6\% on passage-level, 41.6\% on global reasoning. No per-context-length curve is reported. The decay axis is the number of evidence pieces that must be integrated, a $k$-axis quantity rather than an $n$-axis one. NoCha is therefore not counter-evidence to a sublinear-in-$n$ claim. Evidence-scope decay is exactly the $k$-axis observation the framework predicts cannot be addressed by scaling raw context length; it requires retrieval or process supervision along the actual decay axis. The relocation is informational, not magical.

\paragraph{Unifying observation.}
Each of the steep-decay counter-papers we re-audit decays over a variable other than $n$: compositional graph size, fact count, log-time horizon, capacity threshold, or evidence scope. The apparent rapid decay is, in each case, in a quantity our framework already concentrates the action in ($k_{\text{hard}}$ and $|C|$), not in raw sequence length. This is a relocation, not a dissolution. Where $k_{\text{hard}}$ grows with task length (adversarial compositional structure, multi-hop fact chains, long horizons that force more decisions), reliability remains hard; the framework's value is directing intervention toward capability provisioning along the actual decay axis rather than toward context-window or compute-budget expansion that does not help.

\section{Patch Evidence Detail}
\label{app:patch-evidence}

The body's claim that domain patches cap the engineering problem rests on a triangulation of indirect evidence. None of the items below \emph{measures} the patch ceiling $|C_D|$ directly; they support the weaker claim that model behaviour is strongly domain-dependent and that operational neighbourhoods occupy bounded regions of the model's behavioural space.

\begin{center}
\small
\begin{tabular}{p{4.0cm}p{4.5cm}p{4.5cm}}
\toprule
\textbf{Evidence type} & \textbf{What it supports} & \textbf{What it does not prove} \\
\midrule
Low intrinsic-dimensional structure in representations~\citep{park2024linear,li2025stratified} & Domains occupy localised structure in model representation space & Does not measure failure-catalogue size $|C_D|$ \\
\midrule
Cross-domain accuracy spreads on a single base model~\citep{wangy2024mmlupro} & Same model behaves differently across domains; performance is patch-dependent & Does not imply finite failure modes \\
\midrule
Long-tail / popularity thresholds in knowledge tasks~\citep{mallen2023popqa,kandpal2023longtail} & Coverage depends on domain frequency and training exposure & Does not prove logarithmic mode discovery \\
\midrule
Patch-specific intervention performance (across \S\ref{sec:evidence}, Appendix~\ref{app:taxonomy}) & Local tools, schemas, and capability libraries change residual error structure & Does not imply universal cross-patch transfer of the same intervention library \\
\bottomrule
\end{tabular}
\end{center}

The purpose of this evidence is motivational. It supports patch-indexing of $\sigma_D$, $A_D$, $\beta_D$, $|C_D|$, but the finite (or effectively capped) reachable catalogue remains an empirical modelling assumption to be measured per deployment, not derived from any of the rows above.

\section{Empirical Calibration Detail}
\label{app:calibration}

Three published LLM error taxonomies anchor the mode-rate parameter $\sigma$ in the body's $\sigma \in [0.87, 1.85]$ range. The simple calibration uses $|C| \approx A + \sigma \ln T$ with $A = 0$; this is a deliberately conservative readout that ignores any positive intercept and uses endpoint counts rather than discovery curves.

\begin{center}
\small
\begin{tabular}{p{3.5cm}p{2.6cm}p{1.8cm}p{2.4cm}p{3.2cm}}
\toprule
\textbf{Source} & \textbf{Approx.\ observed failures $T$} & \textbf{Named categories $|C|$} & \textbf{Implied $\sigma$ at $A=0$} & \textbf{Role} \\
\midrule
ErrorAtlas~\citep{ashurytahan2026erroratlas} & $\gtrsim 10^4$ & 17 & $\teApprox 1.85$ & Conservative cross-domain anchor (used as planning value) \\
\midrule
HumanEval categorisation~\citep{wen2024fixing} & comparable scale & $8$--$12$ & $\teApprox 0.87$--$1.30$ & Code-domain anchor \\
\midrule
MWPES-300K~\citep{sun2025mwpes} & $\teApprox 3 \times 10^5$ & $15$--$20$ & $\teApprox 1.2$--$1.6$ & Math-domain anchor (largest corpus) \\
\bottomrule
\end{tabular}
\end{center}

These are endpoint counts, not discovery curves: they tell us \emph{how many} named categories a taxonomer assigned at a single corpus scale, not how the count grew with $T$. They do not prove logarithmic mode discovery. Their role in this paper is twofold: they motivate the empirical postulate of \S\ref{sec:postulate}, and they identify the explicit empirical test that would either tighten the postulate or move the analysis to the Heaps variant of Appendix~\ref{app:heaps}: repeatedly subsample failures from a fixed deployment patch $D$ and plot discovered modes $|C_{\text{seen},D}(T)|$ against $T$.


\end{document}